\documentclass{article}
\usepackage{ijcai25}
\usepackage[normalem]{ulem}
\usepackage{enumitem}
\usepackage{times}
\usepackage{soul}
\usepackage{url}
\usepackage[hidelinks]{hyperref}
\usepackage[utf8]{inputenc}
\usepackage[small]{caption}
\usepackage{graphicx}
\usepackage{amsmath}
\usepackage{amsthm}
\usepackage{booktabs}
\usepackage{algorithm}
\usepackage{algorithmic}
\usepackage[switch]{lineno}
\usepackage{mathtools}
\usepackage{threeparttable}
\usepackage{amsfonts,amssymb} 
\usepackage{tabularx}
\usepackage{multirow}
\usepackage{booktabs}
\usepackage{array}
\usepackage{xcolor}
\usepackage{subfigure}
\newcolumntype{C}{>{\centering\arraybackslash}X} 
\newcommand{\enyan}[1]{\textcolor{orange}{(Enyan: #1)}}
\newcommand{\zyx}[1]{\textcolor{blue}{(zyx: #1)}}
\newcommand{\std}[1]{{\scriptsize #1}}

\newtheorem{definition}{Definition}
\newtheorem{problem}{Problem}

\title{SCL-GNN: Towards Generalizable Graph Neural Networks via Spurious Correlation Learning}

\author{
Yuxiang Zhang$^1$\footnote{This work is done during the internship}
\and
Enyan Dai$^1$\footnote{Corresponding author}\\
\affiliations
$^1$The Hong Kong University of Science and Technology (Guangzhou)\\
\emails
\{yuxiangzhang, enyandai\}@hkust-gz.edu.cn
}

\begin{document}

\maketitle


\begin{abstract}
Graph Neural Networks (GNNs) have demonstrated remarkable success across diverse tasks. However, their generalization capability is often hindered by spurious correlations between node features and labels in the graph. Our analysis reveals that GNNs tend to exploit imperceptible statistical correlations in training data, even when such correlations are unreliable for prediction. To address this challenge, we propose the Spurious Correlation Learning Graph Neural Network (SCL-GNN), a novel framework designed to enhance generalization on both Independent and Identically Distributed (IID) and Out-of-Distribution (OOD) graphs. SCL-GNN incorporates a principled spurious correlation learning mechanism, leveraging the Hilbert-Schmidt Independence Criterion (HSIC) to quantify correlations between node representations and class scores. This enables the model to identify and mitigate irrelevant but influential spurious correlations effectively. Additionally, we introduce an efficient bi-level optimization strategy to jointly optimize modules and GNN parameters, preventing overfitting. Extensive experiments on real-world and synthetic datasets demonstrate that SCL-GNN consistently outperforms state-of-the-art baselines under various distribution shifts, highlighting its robustness and generalization capabilities.
\end{abstract}

\section{Introduction}
Graph Neural Networks (GNNs)~\cite{kipf2016semi,velivckovic2017graph} have emerged as powerful models for learning from graph data across various tasks~\cite{bongini2021molecular,cheng2022financial,ni2024graph}. The standard learning paradigm for GNNs involves optimizing parameters on training graphs to make predictions on unseen testing graphs. 
With the increasing need to handle graphs with diverse sources, improving the generalization of GNNs has become a central focus in the graph learning community.

However, recent studies indicate that spurious correlations are prevalent in machine learning~\cite{izmailov2022feature,zhao2024twist,ye2024spurious}, often compromising the generalization of deep neural networks~\cite{degrave2021ai,kirichenko2022last}. 
In GNNs, the issues are particularly severe due to the complex non-Euclidean structure of graph data and the message-passing mechanism. 
Furthermore, unlike standard data modalities (e.g., text and images) where samples are typically treated independently~\cite{wang2023reducing,ghosal2024vision}, the intricate inter-dependencies within graph data (e.g., dense node clusters) significantly increase the difficulty of identifying and mitigating spurious correlations~\cite{wu2022handling,wu2024learning}.


\begin{figure*}[t]
    \centering
    \includegraphics[width=\textwidth]{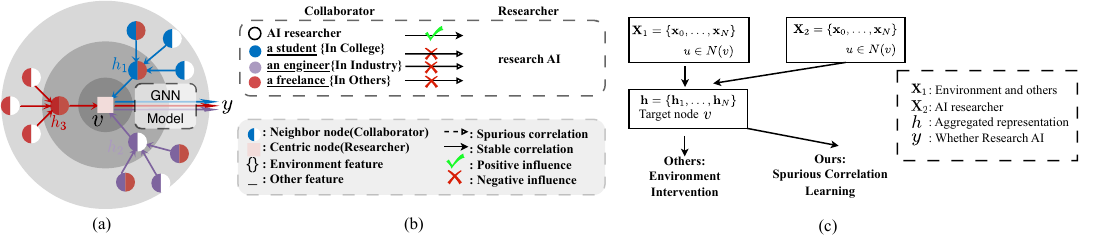}
    \caption{An illustration of node classification task on an academic network for GNNs. (a) The task aims to classify the label $y$ of target node $v$ based on the compact graph $\mathcal{G}_v = \{\mathbf{A}_v, \mathbf{X}_v$\} consisting of its neighbors. (b) Four existing relations and the notion in the network example, only the \textit{AI researcher} feature of the collaborator has a positive influence for predicting that the researcher also researches AI. (c) Description and comparison of the workflow of SCL-GNN and existing works. We virtually split features into spurious and stable sets. We propose a more effective method to achieve the same goal by spurious correlation learning, instead of importing complex probability computation.}
    \label{fig:1}
\end{figure*}

Recent studies show that GNN generalizability degrades primarily because models learn correlations between irrelevant representations and category labels~\cite{chen2022learning,fan2023generalizing,chen2024learning,ye2024spurious}. 
However, existing solutions mainly target Out-of-Distribution (OOD) generalization. 
They often ignore the spurious correlations present in Independent and Identically Distributed (IID) scenarios, which limits their practical applicability. 
Consider a node classification task in an academic network in Fig.\ref{fig:1}, where nodes are researchers, edges are collaborations, and the target label $y$ is whether a researcher specializes in AI. 
GNNs typically rely on stable correlations, such as the actual relationship between a researcher's collaborators and their own field~\cite{wu2024graph}.
Yet, they also absorb spurious correlations, such as a coincidental link between being a student and studying AI. 
In an IID setting, an industry researcher collaborating with both an AI expert and an engineer might be misclassified simply because they lack the "student" attribute. 
This problem worsens under OOD conditions where these spurious correlations shift or disappear entirely (e.g., observing no students in industry settings), leading to systematic mispredictions as detailed in Sec.\ref{sec5.2}. 
Consequently, there is a clear need for GNNs equipped to identify and mitigate spurious correlations across both IID and OOD settings.

Nevertheless, a significant challenge lies in identifying and mitigating the generation of spurious correlations from the broader set of statistical correlations, which are inherently intertwined with the learning mechanisms of Graph Neural Network (GNN) models. To illustrate this, we present an example with Fig.\ref{fig:1}: the marginal probabilities for \textit{a researcher's collaborators have an interest in researching AI}, i.e.~stable correlation, and \textit{a researcher's collaborator is a student}, i.e.~spurious correlation, are both high for the GNN model's prediction of whether \textit{a researcher is researching AI}. Additionally, the scarcity of labeled training data, compounded by the complex non-Euclidean geometric structure of graph data, further complicates the learning process. 
These factors collectively hinder the identification of spurious correlations, which is crucial for ensuring that GNN models avoid relying on non-generalizable patterns in graph learning. Although prior studies~\cite{fan2023generalizing,lin2024graph,wu2024graph} have attempted to address OOD degradation of GNN, exploration of spurious correlation learning on both in-distribution (ID) and OOD samples remains nascent. This is primarily due to the non-trivial nature of the problem, which presents two key challenges: (i) How can spurious correlations be disentangled and distinguished from the broader set of statistical correlations between representations and labels? and (ii) How can sufficient and meaningful information be provided to enhance GNN learning across both ID and OOD samples, particularly in the presence of diverse distribution shifts?



Motivated by the aforementioned challenges, we propose a novel \underline{S}purious \underline{C}orrelation \underline{L}earning GNN(SCL-GNN) framework to reduce the degradation of GNNs caused by the spurious correlation and enhance the generalizability of models. 
SCL-GNN incorporates an intuitive yet theoretically grounded method to determine whether a correlation between a feature and the predicted label is spurious. To achieve this, we employ the Hilbert-Schmidt Independence Criterion (HSIC)~\cite{gohberg1990hilbert} to quantify the independence between node representations and the class scores predicted by the model. Additionally, we utilize Gradient-weighted Class Activation Mapping (Grad-CAM)~\cite{selvaraju2017grad,pope2019explainability} to assess the importance of node representations in influencing the expected class scores, serving as a cross-validation mechanism. These two metrics collectively measure the precise correlation between specific features and label predictions. 
To address spurious correlation, a novel learning objective is deployed to fine-tune the model weights through a collaborative self-supervised learning module. Our approach outperforms existing methods in decorrelating spurious relationships and significantly enhances the model's ability to generalize across domains. 


To summarize, our main contributions are: (i) We study a novel problem of addressing the degradation of GNNs under distribution shift from a spurious correlation learning perspective empirically and theoretically. We ensure that GNNs trained with this learning purpose would capture the pattern of spurious correlations and mitigate the deficiency. (ii) We proposed a novel framework that can effectively learn spurious correlation and perform better under complex distribution shifts. The model in the framework utilizes adequate and rich information to avoid overfitting using the auxiliary self-supervised spurious correlation learner module that fine-tunes the model with the unobserved OOD samples. (iii) We conduct comprehensive experiments on various datasets with different distributions. The convincing results show that the SCL-GNN framework significantly outperforms competitors regarding effectiveness, flexibility, and interpretability. 

\section{Related Works}
\textbf{Graph Neural Networks}.
GNNs have shown remarkable success in learning graph-structured data~\cite{kipf2016semi}, driving progress in applications like drug discovery~\cite{bongini2021molecular}, financial analysis~\cite{cheng2022financial}, and recommendation systems~\cite{ying2018graph}. However, their susceptibility to spurious correlations limits broader applicability~\cite{karczewski2024generalization}. Even under the ideal IID assumption, where training and testing graphs share the same distribution, GNNs often suffer performance degradation due to spurious correlations~\cite{eastwood2024spuriosity}. Despite its importance, improving GNN generalizability through spurious correlation learning remains underexplored~\cite{hu2020open,koh2021wilds}. In this work, we address this gap by identifying the key to generalizable GNNs: learning and mitigating spurious correlations. We provide a theoretical analysis on this and propose a novel framework for advance research.

\noindent\textbf{Correlation Learning on Graphs}.
Spurious correlations are a primary cause of OOD performance degradation~\cite{ming2022impact,zhang2023robustness}. Prior work has explored decorrelation methods for graph-based models to eliminate dependencies between relevant and irrelevant representations across distributions~\cite{kuang2020stable,li2022ood,fan2023generalizing}. However, these methods often reduce effective sample size, impairing GNN generalization~\cite{zhang2023map,chen2024learning}. Alternative approaches leverage invariant principles~\cite{wu2022handling,chen2022ba} or causal intervention~\cite{chen2022learning,wu2024graph} to identify patterns consistent across domains. Although effective in distribution shifts, causal inference requires a top-down understanding of data generation mechanisms~\cite{tang2023towards,wu2024graph}, involving complex modeling of causal relationships between labels, node features, and representations~\cite{wu2024learning}, often complicated by confounding variables or reverse causation~\cite{chen2022learning,gui2024joint}. Notably, spurious correlations also degrade GNN performance under IID settings, a point underexplored in prior work. We propose a novel framework leveraging spurious correlation learning to enhance generalization under both IID and OOD settings, employing standard machine learning techniques for intuitive and theoretically feasible solutions.



\section{Problem Formulation}\label{sec:problem}

\textbf{Notations}. We denote an attributed graph $\mathcal{G} = (\mathcal{V}, \mathcal{E}, \mathbf{X})$, where $\mathcal{V} = \{v_1, \dots, v_N\}$ is the set of $N$ nodes, $\mathcal{E} \subseteq \mathcal{V} \times \mathcal{V}$ represents the set of edges, and $\mathbf{X} = \{\mathbf{x_1}, \dots, \mathbf{x_N}\}$ indicates node features. The adjacency matrix of the graph $\mathcal{G}$ is denoted as $\mathbf{A} \in \mathbb{R}^{N \times N}$, where $A_{ij}=1$ if nodes $v_i$ and $v_j$ are connected. Given the training graphs $\mathcal{G}_{train}$, the validation graphs $\mathcal{G}_{val}$ and the testing graphs $\mathcal{G}_{test}$. ${p}(\mathcal{G}_{train}) \neq {p}(\mathcal{G}_{val}) \neq {p}(\mathcal{G}_{test})$ denotes unknown distribution shift, otherwise ${p}(\mathcal{G}_{train}) = {p}(\mathcal{G}_{val}) = {p}(\mathcal{G}_{test})$. $Z = f(\mathcal{G})$, $Z \in \mathbb{R}^{N \times d}$ is the output of the GNN model $f(\cdot)$.

\noindent \textbf{Formulating Spurious Correlation}.
Spurious correlation, namely ``\textit{correlations that do not imply causation}" in statistics, refers to a situation where two variables appear to be related to each other, yet the relationship is coincidental or confounded by an external variable~\cite{ye2024spurious}. The correlation could lead to incorrect interpretations of models. Spurious correlation in graph learning can be defined as the potential erroneous relationship learned by GNNs between the specific irrelevant sets of node representations and labels of nodes, which can be formalized as
\begin{definition}\label{def:sc}
    Let $\mathcal{G}_{trian}=\{(v_i,y_i)\}_{i=1}^n$ be the training graph with $v_i\in\mathcal{V}$ and $y_i\in\mathcal{Y}$, where $\mathcal{V}$ denotes the set of all training nodes, ${\mathcal Y}$ denotes the set of $K$ classes. For each node $v_i$ with label $y_i$, there is the spurious feature $a_i \in \mathcal{A}$ of $v_i$, where $a_i$ cannot help the model make a correct prediction of $y_i$, and $\mathcal{A}$ denotes the set of all possible spurious features. 
    A spurious correlation, denoted as $\langle y,a\rangle$, is the association between $y\in\mathcal{Y}$ and $a\in\mathcal{A}$, where $y$ and $a$ exist in a one-to-many mapping as:$\phi:\mathcal{A}\mapsto\mathcal{Y}^{K'}$ conditioned on $\mathcal{G}_{trian}$ with $1<K'\leq K$. 
\end{definition}

\vspace{0.3em}
\noindent \textbf{Mitigate Spurious Correlation}. 
In the context of generalizing on graphs, spurious correlations pose significant threats, as features that are predictive in the training distribution ${p}(\mathcal{G}_{train})$ may no longer correlate with the label in the testing distribution ${p}(\mathcal{G}_{test})$. Notably, the presence of a distribution shift, that is, OOD, is not a prerequisite~\cite{ye2024spurious}. To mitigate spurious correlations, we fine-tune the weight matrix $\mathbf{W}$ of the surrogate GNN model $f_s$ to reduce the influence of spurious features on the model predictions.

In particular, since the labeled data available during model training is limited, overfitting can easily occur, reducing the quality of the learned spurious correlations. A standard approach to address the scarcity of labeled data is to adapt the trained model to the unlabeled distribution before making predictions through a self-supervised auxiliary task~\cite{xie2022self}. To address this, we introduce self-supervise learning method in spurious correlation learner module, which will be detailed in Sec.\ref{sec:scl}. 
With all the above descriptions and notations, we can formulate the problem of generalizing GNN by learning the spurious correlation as:

\begin{problem}\label{problem1}

    Given an attributed graph $\mathcal{G} = (\mathcal{V}, \mathcal{E}, \mathbf{X})$. We aim to learn an adaptive spurious correlation learner $f_a: \mathbf{W} \xrightarrow{\text{\sout{$\langle y,a\rangle$}}} \mathbf{W}'$ to fine-tune the weight matrix $\mathbf{W}$ of the trained GNN model $f_\theta$ and mitigate the impact of spurious correlations by solving:
    \begin{equation}\label{eq:bilevel}
    \begin{aligned}
        &\min_{\theta_{a}} \sum_{v_i \in \mathcal{V}} l(f_{\theta^*}(\mathcal{G}), y_i) + \beta\mathcal{L}_{\mathcal{S}}(\theta^*, \theta_a) \\
        &s.t.\theta^* = \mathop{\arg\min}_{\theta} \sum_{v_i \in \mathcal{V}_L}l(f_{\theta}(\mathcal{G}_{train}), y_i) + \gamma\mathcal{R}(\theta, \theta_a) \\
    \end{aligned}
    \end{equation}
    where $l(\cdot)$ denotes the cross-entropy loss, $\mathcal{L_{S}}$ denotes spurious correlation learning loss, and $\beta$ is the hyperparameter to control the contribution of $\mathcal{L_{S}}$. $\theta_a$ denotes the parameters of $f_a$, and $\theta^*$ indicates the trained parameters of the GNN model. $\mathcal{V}_L$ indicates the labeled nodes, and $\mathcal{R}(\theta, \theta_a)$ is a regularization constraint ties $\theta$ and $\theta_a$.
\end{problem}

\begin{figure*}[ht]
    \centering
    \includegraphics[width=\textwidth]{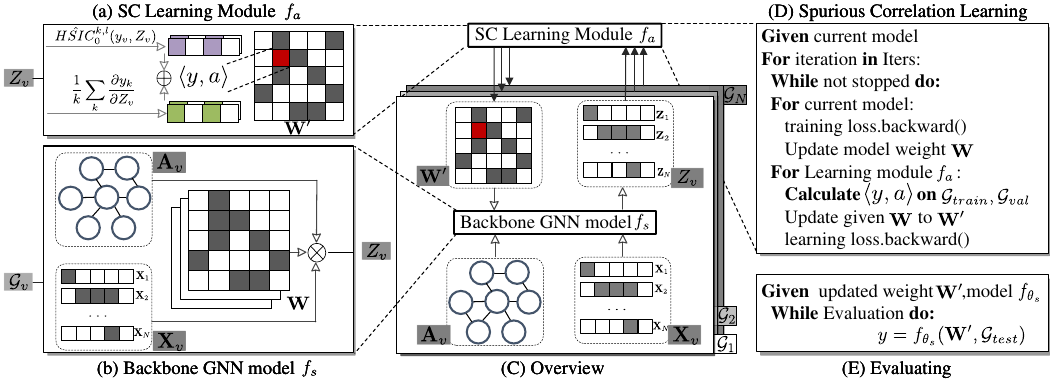}
    \caption{Illustration of SCL-GNN framework for generalizing GNN via spurious correlation learning. (a) and (b) illustrates the details of spurious correlation(abbreviated as SC) learner $f_a$ and backbone GNN model $f_s$, respectively, (c) represent an overview of the framework, (d) and (e) respectively explain the process of spurious correlation learning and generalization via pseudo algorithms. When the training is finished, the $f_a$ is fixed synchronously, and we only need to evaluate GNN model with the fine-tuned model weight $\mathbf{W}^\prime$.}
    \label{fig:2}
\end{figure*}
\section{Methodology}
In this section, we present details of SCL-GNN framework which optimizes Eq.(\ref{problem1}) to learn and mitigate the spurious correlation between feature and label. As it is challenging to directly optimize the Eq.(\ref{problem1}), SCL-GNN separately optimizes the spurious correlation learner $f_a$ and the surrogate GNN model $f_s$, and we introduce Pareto efficiency for the multi-objective optimization~\cite{sener2018multi}. We illustrate the overview of SCL-GNN in Fig~\ref{fig:2}.
\subsection{Learning Module}\label{sec:scl}
To ensure the effectiveness of the learning module \( f_a \), we propose a differentiable non-linear spurious correlation learning loss, defined as:
\begin{equation}\label{eq:scl}
    \mathcal{L_{S}} =  \sum\limits_{{v}\in \mathcal{V}_{S}}l_{scl}(\langle y, Z_v\rangle,(\theta_s, \theta_a)),
\end{equation}
where $\mathcal{L_{S}}$ denotes the loss function of the learning module, $\mathcal{V}_{S} = \mathcal{V}\setminus\mathcal{V}_{test}$ represents the set of labeled and unlabeled nodes. $l_{scl}(\cdot)$ is the spurious correlation learning loss, and $\theta_s$ represents the shared parameters between the current model and the learning module. The learner $\theta_a$ assists the surrogate model $f_s$ in avoiding spurious features and mitigating spurious correlations through the optimization of $\mathcal{L_{S}}$.
Our approach to solving Problem~\ref{problem1} centers on optimizing a margin loss formulated as the difference between two terms: the Hilbert-Schmidt Independence Criterion (HSIC) and Gradient-weighted Class Activation Mapping (Grad-CAM). Specifically, a larger HSIC term and a smaller Grad-CAM term indicate a stronger spurious correlation between the node feature and its prediction. The formulation provides a principled mechanism to quantify and reduce spurious correlations for the generalizability.

\subsubsection{Irrelevance Measurement} Nonlinear dependency between two variables can be measured using the Hilbert-Schmidt Independence Criterion (HSIC). According to the previous article's definition of node-level spurious correlation, this correlation exists between the class prediction of a node and its representation. We first use the improved HSIC algorithm to calculate the score between the two representations. For two random variables $U$ and $V$ and kernel functions $k$ and $l$, the corresponding HSIC is defined as:
\begin{equation}\label{eq:hsic}
    HSIC^{k,l}(U,V) := ||C^{k,l}_{U,V}||^{2}_{h},
\end{equation}
where $C^{k,l}$ indicates the cross-covariance operator in kernel function's Reproducing Kernel Hilbert Spaces (RKHS) $k$ and $l$~\cite{van2008reproducing}. $||\cdot||_{h}$ indicates the Hilbert-Schmidt norm. $HSIC^{k,l}(U,V) = 0$ if and only if $U \perp V$, i.e, the variable $U$ and $V$ is totally irrelevant and $k$, $l$ are both Gaussian kernels. We further adapt a widely used estimator~\cite{gretton2005measuring} with $m$ samples to adapt the lack of labeled nodes, defined as:
\begin{equation}\label{eq:hsic}
    HSIC_0^{k,l}(U,V) = (m-1)^{-2}\mathbf{tr}\mathbf{(UPVP)},
\end{equation}
where $\mathbf{tr}(\cdot)$ indicates the trace function, and $\mathbf{U},\mathbf{V} \in \mathbb{R}^{k\times k}$ are Gaussian kernel matrices containing entities $\mathbf{U}_{ij} = k(U_i, U_j)$ and $\mathbf{V}_{ij} = l(V_i, V_j)$. $\mathbf{P} = \mathbf{I} - m^{-1}\mathbf{1}\mathbf{1}^{T}\in \mathbb{R}^{k\times k}$ is the centering matrix, where $\mathbf{I}$ is an identity matrix and $\mathbf{1}$ is a all-one column. 

To establish the HSIC relevance between the score $y_{v_i}$ and output ${Z}_{v_i}$ of node $v_i$, the Eq.(\ref{eq:hsic}) can be reformulated as:
\begin{equation}
\begin{gathered}
    H\hat{SI}C_0^{k,l}(y_{v_i},{Z}_{v}) := \sigma(||C^{~~k,~~l}_{y_{v_i},{Z}_{v}}||^{2}_{F})\\
    =\sigma((m-1)^{-2}\mathbf{tr}{({y}_{v_i}\mathbf{P}{Z}_v)\mathbf{P})}),
\end{gathered}
\end{equation}
where \(||\cdot||^2_{F}\) denotes the Frobenius norm. When the operator in the Hilbert space is finite-dimensional, \(||\cdot||^2_{F}\) is equivalent to \(||\cdot||_{HS}^2\), a property we leverage to reduce computational costs. Additionally, \(\sigma\) represents a normalization operator. We normalize the computed results to the interval \([0, 1]\) to simplify subsequent calculations.
\subsubsection{Importance Measurement} Generally, An intuitive explanation for GNN models is identifying node features that strongly influence final predictions~\cite{pope2019explainability}. To demonstrate which feature plays an important role while the current model is conducting prediction, we use the Gradient-weighted Class Activation Mapping (Grad-CAM) based explanation method~\cite{selvaraju2017grad} to calculate the importance of the various dimensions of node feature concerning the class score given by the model. We present the computing process as follows:
\begin{equation}\label{eq:scorek}
    y_{k,v} = \frac{w_k}{|\mathcal{G}_v|}\sigma\left[V W^L H^{L-1}(\mathbf{A}_v, \mathbf{X}_v)\right]
\end{equation}
where $y_{k,v}$ is the score for the $k$-th dimension of the feature of node $v$, $w_k$ are the weights learned based on the input-output behavior of the current model, $\sigma(\cdot)$ is the element-wise nonlinear activation function $W^L$ is the weight of the last layer $L$ of the model, and $H(\cdot)$ is the convolutional activation. $V$ indicates ${{\widetilde D}^{ - \frac{1}{2}}}\widetilde A{{\widetilde D}^{ - \frac{1}{2}}}$, where $\widetilde A = A + I_N$ is the adjacency matrix with added self-loops and $I_N\in \mathbb{R}^{N\times N}$ is the identity matrix, $\widetilde D_{ii} = \sum_{j}\widetilde A_{ij}$, and $\widetilde A = A + I_N$. After obtaining the score representation, we represent the importance of the node representation to the model prediction as:
\begin{equation}
    i_{v} = \frac{1}{k}\sum_{k}\frac{{\partial{y_{k}}}}{{\partial{Z_v}}}
\end{equation}
Based on the analysis, Eq.(\ref{eq:scl}) can be rewritten as:
\begin{equation}\label{eq:rewrite}
\begin{aligned}
    &\mathcal{L_{S}}(\theta_{s}, \theta_{a}) = \\
    &\sum\limits_{{v}\in \mathcal{V}_{S}}\max\left[0, (H\hat{SI}C_0^{k,l}(y_{v},{Z}_v)) - \frac{1}{k}\sum_{k}\frac{{\partial{y_{k}}}}{{\partial{Z_v}}})\right]\\
\end{aligned}
\end{equation}
and the Problem~\ref{problem1} can be reformulated as:
\begin{equation}\label{eq:reform}
\begin{aligned}
    &\mathbf{W^*} = \mathop{\arg\min}_{\mathbf{W}} \sum_{v_i \in \mathcal{V}} l(f_{\theta^*}({G}), y_t) + \mathcal{L}_{\mathcal{S}}(\theta^*, \theta_a) \\
    &s.t. \quad \theta^* = \mathop{\arg\min}_{\theta}  \sum_{v_i \in \mathcal{V}_L}l(f_{\theta}({G}_{train}), y_i) + \gamma\mathcal{R}(\theta, \theta_a) ,
\end{aligned}
\end{equation}
where $\mathbf{W}$ indicates the optimized weight matrix of the GNN model. We minimize the model’s reliance on spurious correlation in learning graph representations by jointly optimizing the GNN $\theta_s$, learner $\theta_a$, and weights $\mathbf{W}$.

\subsection{Bi-level optimization}
Inspired by previous work~\cite{chen2024learning}, we chose not to solve the lower-level problem for each outer loop. To guarantee the computing effectiveness, we update the parameter $\theta_s$ and only perform the gradient computation with the learning rate $\eta_{\theta_s}$ that are listed as:
\begin{equation}
    \theta^{(i)}= \theta^{(i-1)} - \eta_{\theta}\nabla_{\theta}\mathcal{L}_{train}(\theta^{(i-1)}),
\end{equation}
where $\mathcal{L}_{train}$ is correspond to the constraint of Eq.(~\ref{eq:bilevel}), $\theta^i$ indicates the model parameter after $i$ iterations.

We compute the upper-level objective in the inner loop after getting the parameter $\theta$ able to indicate model weight $W$, which is an estimate of $\theta^*$. Then we update the parameter $\theta_a$ of $f_a$ by applying first-order approximation~\cite{finn2017model} in computing gradients of $\theta_a$ to reduce the computing cost further:
\begin{equation}
    \theta_a^{(t+1)}= \theta_a^{(t)} - \eta_{\theta_a}\nabla_{\theta_a}\mathcal{L}_{\mathcal{S}}(\widehat{\theta_{s}}, \theta_a^{(t)}),
\end{equation}
where $\widehat{\theta_{s}}$ indicates the gradient propagation stopping. $\eta_{\theta_a}$ is the learning rate of learner module $f_a$.

\section{Experiments}
In this section, we describe the experimental setup employed to assess the efficacy of our proposed framework. The results demonstrate our framework's superior performance compared to various baselines on diverse datasets. Specifically, we address the following research questions: 
\begin{itemize}[leftmargin=*]
    \item \textbf{RQ1}: How does SCL-GNN perform relative to state-of-the-art methods in addressing spurious correlations?
    \item \textbf{RQ2}: What is the sensitivity of SCL-GNN w.r.t. the weight of learning loss $\beta$ and correlations $f_a$ gained?
    \item \textbf{RQ3}: Does the proposed bi-level optimization effectively alleviate the targeted issue?
    \item \textbf{RQ4}: Do the proposed methods function as intended and provide meaningful insights?
\end{itemize}
\begin{table}[th]
	\centering
	\caption{Details of Datasets.}
	\label{tab-datasets}
	\begin{threeparttable}
	\setlength{\tabcolsep}{2mm}{
	\resizebox{0.49\textwidth}{!}{
    \begin{tabular}{lccccc}
        \toprule 
        Datasets  & \#Nodes & \#Edges & \#Features & \#Classes  & Distribution Shift \\
        \midrule
   \texttt{Cora} & 2708 & 5429 & 1433 & 7 & Features \\
   \texttt{Pubmed} & 19717 & 44338 & 500 & 3 & Features\\
   \texttt{Arxiv} & 169343 & 1166243 & 128 & 40 & Timeline \\
   \texttt{Products} & 2449029	& 61859140 & 100 & 47 & Popularity \\
        \bottomrule
	\end{tabular}}}
	\end{threeparttable}
\end{table}

\begin{table*}[t]
    \centering
    \small
    \caption{Accuracy $\pm$ Standard Deviation(\%) for \texttt{Cora} and \texttt{Pubmed} on ID and two sets of OOD data, respectively. The distribution shift is generated on feature of nodes.}
    \label{tab-cora}
    \begin{tabularx}{\textwidth}{c*{8}C}
    \toprule
    \multirow{2}{*}{\textbf{Backbone}} & \multirow{2}{*}{\textbf{Method}} & \multicolumn{3}{c}{\texttt{Cora}} & \multicolumn{3}{c}{\texttt{Pubmed} }\\
    \cmidrule(lr){3-5} \cmidrule(lr){6-8}
    & &\textbf{ID} & \textbf{OOD1} & \textbf{OOD2}& \textbf{ID} & \textbf{OOD1} & \textbf{OOD2}\\
    \midrule
    \multirow{5}{*}{GCN}
    & StableGNN & 84.05\std{ $\pm$ 2.16}  & 77.31\std{ $\pm$ 0.17} & 73.61\std{ $\pm$ 0.37} & 71.85\std{ $\pm$ 2.23}& 67.12\std{ $\pm$ 0.76} & 65.73\std{ $\pm$ 1.58} \\
    & SRGNN & 91.57\std{ $\pm$ 2.81} & 80.30\std{ $\pm$ 0.21} & 76.31\std{ $\pm$ 0.41}  & 90.89\std{ $\pm$ 4.23} & 82.45\std{ $\pm$ 0.07} & 79.25\std{ $\pm$ 0.17} \\
    & EERM & 88.57\std{ $\pm$ 1.24} & 83.50\std{ $\pm$ 0.44} & 81.07\std{ $\pm$ 0.25} & 82.79\std{ $\pm$ 3.34} & 77.15\std{ $\pm$ 0.24}  & 74.43\std{ $\pm$ 0.65} \\
    & CANET & 96.25\std{ $\pm$ 2.56} & 95.14\std{ $\pm$ 0.56} & 93.89\std{ $\pm$ 0.16} & 96.13\std{ $\pm$ 3.32} & 90.67\std{ $\pm$ 0.26}  & 88.49\std{ $\pm$ 0.04} \\
    & SCL-GNN & 97.14\std{ $\pm$ 1.26} & 95.62\std{ $\pm$ 0.06} & 95.03\std{ $\pm$ 0.09}  & {97.24}\std{ $\pm$ 0.06} & 94.16\std{ $\pm$ 0.76} & 91.56\std{ $\pm$ 0.06}  \\
    \midrule
    \multirow{5}{*}{GAT}
    & StableGNN & 84.67\std{ $\pm$ 2.40}  & 77.89\std{ $\pm$ 0.30} & 73.45\std{ $\pm$ 0.50} & 72.25\std{ $\pm$ 2.50}& 67.50\std{ $\pm$ 1.00} & 65.95\std{ $\pm$ 1.75} \\
    & SRGNN & 93.20\std{ $\pm$ 3.05} & 81.50\std{ $\pm$ 0.45} & 76.80\std{ $\pm$ 0.60}  & 91.75\std{ $\pm$ 4.50} & 83.35\std{ $\pm$ 0.25} & 80.45\std{ $\pm$ 0.35} \\
    & EERM & 89.85\std{ $\pm$ 1.50} & 84.75\std{ $\pm$ 0.65} & 82.20\std{ $\pm$ 0.40} & 84.10\std{ $\pm$ 3.65} & 78.55\std{ $\pm$ 0.40}  & 75.60\std{ $\pm$ 0.85} \\
    & CANET & 97.35\std{ $\pm$ 2.80} & 96.20\std{ $\pm$ 0.75} & 94.70\std{ $\pm$ 0.30} & 97.40\std{ $\pm$ 3.50} & 91.80\std{ $\pm$ 0.45}  & 89.65\std{ $\pm$ 0.30} \\
    & SCL-GNN & 97.90\std{ $\pm$ 1.40} & 96.75\std{ $\pm$ 0.15} & 95.90\std{ $\pm$ 0.20}  & 98.25\std{ $\pm$ 0.20} & 95.05\std{ $\pm$ 0.90} & 92.80\std{ $\pm$ 0.20}  \\
    \bottomrule
    \end{tabularx}
\end{table*}

\subsection{Experimental Settings}
\subsubsection{Datasets and competitors} We utilize four node classification datasets of varying sizes and domains: \texttt{Cora}, \texttt{Pubmed}~\cite{kipf2016semi}, \texttt{Arxiv}, and \texttt{Products}~\cite{hu2020open}. 
We compare our framework SCL-GNN with representative state-of-the-art methods: SRGNN~\cite{zhu2021shift}, StableGNN~\cite{fan2023generalizing}, EERM~\cite{wu2022handling} and CANET~\cite{wu2024graph}, which are all focusing on learning generalizable GNNs. We adopt GCN and GAT as the backbone of all the methods, respectively, and hyperparameters such as learning rate are kept consistent for fairness. As for competitors are OOD generalization oriented, and distribution shift causes more severe spurious correlation, we modify datasets as follow: for \texttt{Cora} and \texttt{Pubmed}, we synthetically generate two sets of spurious features, following the methodology of~\cite{deshpande2018contextual}, to induce distribution shifts between in-distribution (ID) and OOD subsets. For \texttt{Arxiv}, we define publications from 2005–2016 as ID data, 2016–2018 as OOD1 data, and 2019–2020 as OOD2 data. For \texttt{Products}, we employ sales rankings (popularity) to partition the data, with products in the top 20\% classified as ID data, those ranked 20–60\% as OOD1 data, and the remaining as OOD2 data. Detailed statistics of the datasets are summarized in Table~\ref{tab-datasets}.

\subsubsection{Evaluation Protocol}
In this paper, we conduct experiments on the node classification task. For each dataset, we randomly partition the in-distribution (ID) portion of nodes into training, validation, and test sets with a ratio of $15/15/70\%$. The training set is used to train the model $f_s$, while the validation set is employed for early stopping and hyperparameter tuning. The test set is utilized to evaluate the model's performance. To assess the OOD generalization capabilities of SCL-GNN, we evaluate the model's performance on both the ID test data and the OOD data separately. To mitigate randomness, we report the average results over ten independent runs.

\subsection{Performance Comparison}\label{sec5.2}
To answer \textbf{RQ 1}, we compare SCL-GNN with four competitors on four real-world graph datasets. Tabel~\ref{tab-cora} reports the testing accuracy on \texttt{Cora} and \texttt{Pubmed}, from which We found that using either GCN or GAT as the backbone, SCL-GNN consistently outperforms the corresponding competitors by a significant margin on the OOD data across two types of distribution shifts and datasets. At the same time, SCL-GNN shows highly competitive results on the ID data. Apart from the relative improvement over the competitors, we observed that on these two datasets, the performance of SCL-GNN under the OOD settings is very close to that of ID data. These results show that our model can effectively handle distribution shifts caused by various spurious features. Tabel~\ref{tab-arxiv} reports the testing accuracy on \texttt{Arxiv} and \texttt{Products}. To create distribution shifts, we artificially split \texttt{Arxiv} along the timeline. When the time distribution shift between training nodes and test nodes becomes larger, all methods suffer degradation to some extent. Still, SCL-GNN suffers the most minor performance degradation, which stands for both backbones. Especially on the OOD2 dataset, which is the most difficult for other methods to make correct predictions, we achieve 5.77\% and 7.13\% improvement in accuracy on the testing set on the two backbones compared to the second-best CANET. For \texttt{Products}, we found that degradation becomes more significant as we conduct a more challenging splitting procedure that closely matches the real-world application where labels are first assigned to essential nodes in the network, which forces models subsequently to make predictions on less important ones. This gap is more significantly enlarged when we make OOD portions, so we consider \texttt{Products} as the most challenging dataset for generalization in this paper. The size of \texttt{Products} and the algorithm's complexity have caused competitive competitor EERM OOM errors. Still, we found that SCL-GNN achieves overall superior performance over the competitors, demonstrating our model's efficacy for tackling OOD generalization across real-world graphs in representation learning.

\begin{table*}[t]
    \centering
    \small
    \caption{Accuracy $\pm$ Standard Deviation(\%) for \texttt{Arxiv} and \texttt{Products} on ID and two sets of OOD data, respectively. The distribution shift is generated by the timeline and popularity of nodes.}
    \label{tab-arxiv}
    \begin{tabularx}{\textwidth}{c*{8}C}
    \toprule
    \multirow{2}{*}{\textbf{Backbone}} & \multirow{2}{*}{\textbf{Method}} & \multicolumn{3}{c}{\texttt{Arxiv}} & \multicolumn{3}{c}{\texttt{Products} }\\
    \cmidrule(lr){3-5} \cmidrule(lr){6-8}
    & &\textbf{ID} & \textbf{OOD1} & \textbf{OOD2}& \textbf{ID} & \textbf{OOD1} & \textbf{OOD2}\\
    \midrule
    \multirow{5}{*}{GCN}
    & StableGNN & 60.35\std{ $\pm$ 0.46}  & 57.33\std{ $\pm$ 0.12} & 53.25\std{ $\pm$ 0.49} & 80.23\std{ $\pm$ 0.43}& 69.12\std{ $\pm$ 0.46} & 67.23\std{ $\pm$ 0.38} \\
    & SRGNN & 61.02\std{ $\pm$ 0.18} & 57.30\std{ $\pm$ 0.11} & 46.98\std{ $\pm$ 0.29}  & 84.16\std{ $\pm$ 0.42} & 78.12\std{ $\pm$ 0.07} & 72.56\std{ $\pm$ 0.13} \\
    & EERM & 57.12\std{ $\pm$ 0.14} & 53.15\std{ $\pm$ 0.44} & 48.37\std{ $\pm$ 0.19} & 82.42\std{ $\pm$ 0.42} & 74.52\std{ $\pm$ 0.33}  & 72.45\std{ $\pm$ 0.17} \\
    & CANET & 62.52\std{ $\pm$ 0.20} & 59.04\std{ $\pm$ 0.22} & 56.12\std{ $\pm$ 0.15} & 85.61\std{ $\pm$ 0.33} & 78.27\std{ $\pm$ 0.46}  & 75.51\std{ $\pm$ 0.19} \\
    & SCL-GNN & 63.74\std{ $\pm$ 0.36} & 62.15\std{ $\pm$ 0.06} & 60.89\std{ $\pm$ 0.22}  & 88.04\std{ $\pm$ 0.15} & 84.16\std{ $\pm$ 0.09} & 80.07\std{ $\pm$ 0.20} \\
    \midrule
    \multirow{5}{*}{GAT}
    & StableGNN & 60.79\std{ $\pm$ 1.01}  & 56.24\std{ $\pm$ 0.16} & 53.95\std{ $\pm$ 0.26} & OOM  & OOM & OOM\\
    & SRGNN & 61.39\std{ $\pm$ 0.17} & 57.23\std{ $\pm$ 0.41} & 51.27\std{ $\pm$ 0.33}  & OOM & OOM  & OOM  \\
    & EERM & 58.62\std{ $\pm$ 0.34} & 54.32\std{ $\pm$ 0.24} & 49.71\std{ $\pm$ 0.37} & 82.68\std{ $\pm$ 0.46}& 76.52\std{ $\pm$ 0.40} & 71.33\std{ $\pm$ 0.81}\\
    & CANET & 64.01\std{ $\pm$ 0.41} & 59.22\std{ $\pm$ 0.37} & 54.61\std{ $\pm$ 0.26} & 84.11\std{ $\pm$ 0.36} & 79.21\std{ $\pm$ 0.16}  & 77.37\std{ $\pm$ 0.13} \\
    & SCL-GNN & 64.22\std{ $\pm$ 0.46} & 61.48\std{ $\pm$ 0.33} & 59.37\std{ $\pm$ 0.17}  & 85.69\std{ $\pm$ 0.42} & 83.92\std{ $\pm$ 0.22} & 81.44\std{ $\pm$ 0.33} \\
    \bottomrule
    \end{tabularx}
\end{table*}

\subsection{Sensitivity and Ablation Studies}
\begin{figure}[ht]
\centering
\begin{minipage}[t]{\linewidth}
\subfigure[GCN-\texttt{Cora}]{
\begin{minipage}[t]{0.5\linewidth}
\centering
\includegraphics[width=\textwidth]{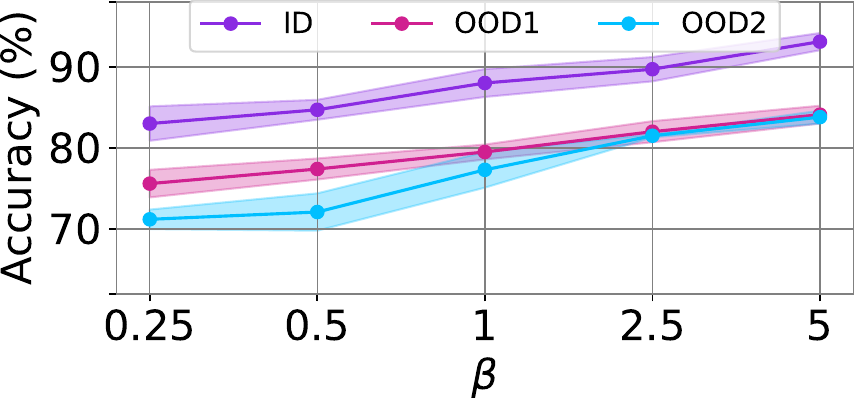}
\end{minipage}%
}%
\subfigure[GAT-\texttt{Cora}]{
\begin{minipage}[t]{0.5\linewidth}
\centering
\includegraphics[width=\textwidth]{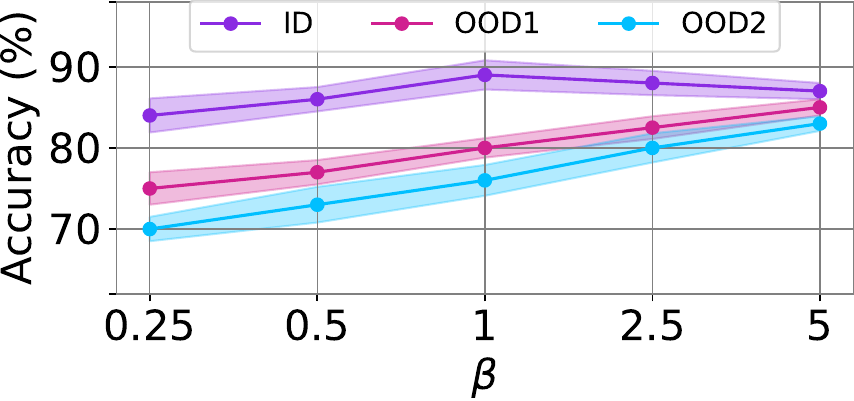}
\end{minipage}%
}%

\subfigure[GCN-\texttt{Arxiv}]{
\begin{minipage}[t]{0.5\linewidth}
\centering
\includegraphics[width=\textwidth]{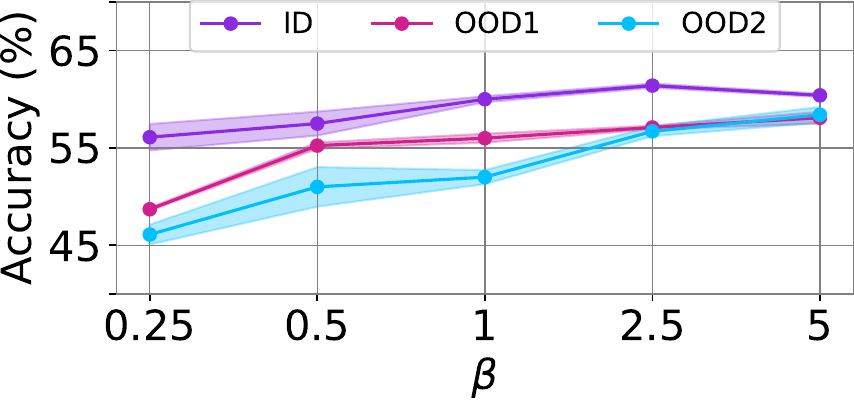}
\end{minipage}%
}%
\subfigure[GAT-\texttt{Arxiv}]{
\begin{minipage}[t]{0.5\linewidth}
\centering
\includegraphics[width=\textwidth]{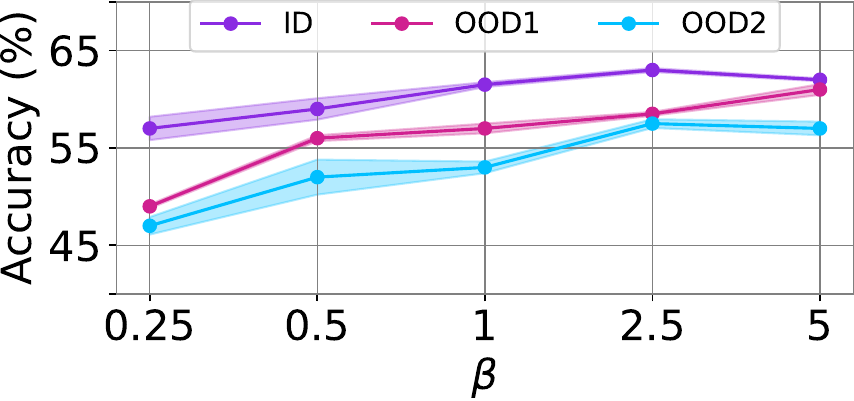}
\end{minipage}%
}%
\end{minipage}
\caption{Model performance with different $\beta$.}
\label{fig:3}
\end{figure}

\noindent\textbf{Sensitivity Study on Hyperparameter}
To answer \textbf{RQ 2}, we vary the hyperparameter $\beta$, which controls the weight of the spurious correlation learning loss, across the values $\{0.25, 0.5, 1, 2.5, 5\}$. Figure~\ref{fig:3} illustrates the performance of backbone models (GCN and GAT) on the \texttt{Cora} and \texttt{Arxiv} datasets under different $\beta$ values. As $\beta$ increases, both GCN and GAT exhibit improved performance. However, three key observations emerge: (i) While increasing $\beta$ enhances performance by mitigating spurious correlations, excessive values lead to underfitting and performance degradation. (ii) At optimal $\beta$ values, the model achieves comparable performance on both in-distribution (ID) and out-of-distribution (OOD) samples, indicating enhanced generalization. (iii) Despite these improvements, the results in Tables~\ref{tab-cora} and~\ref{tab-arxiv} reveal a significant performance gap, underscoring the critical role of the learner in boosting the model's generalization capabilities.

\begin{figure}[ht]
    \centering
    \includegraphics[width=\linewidth]{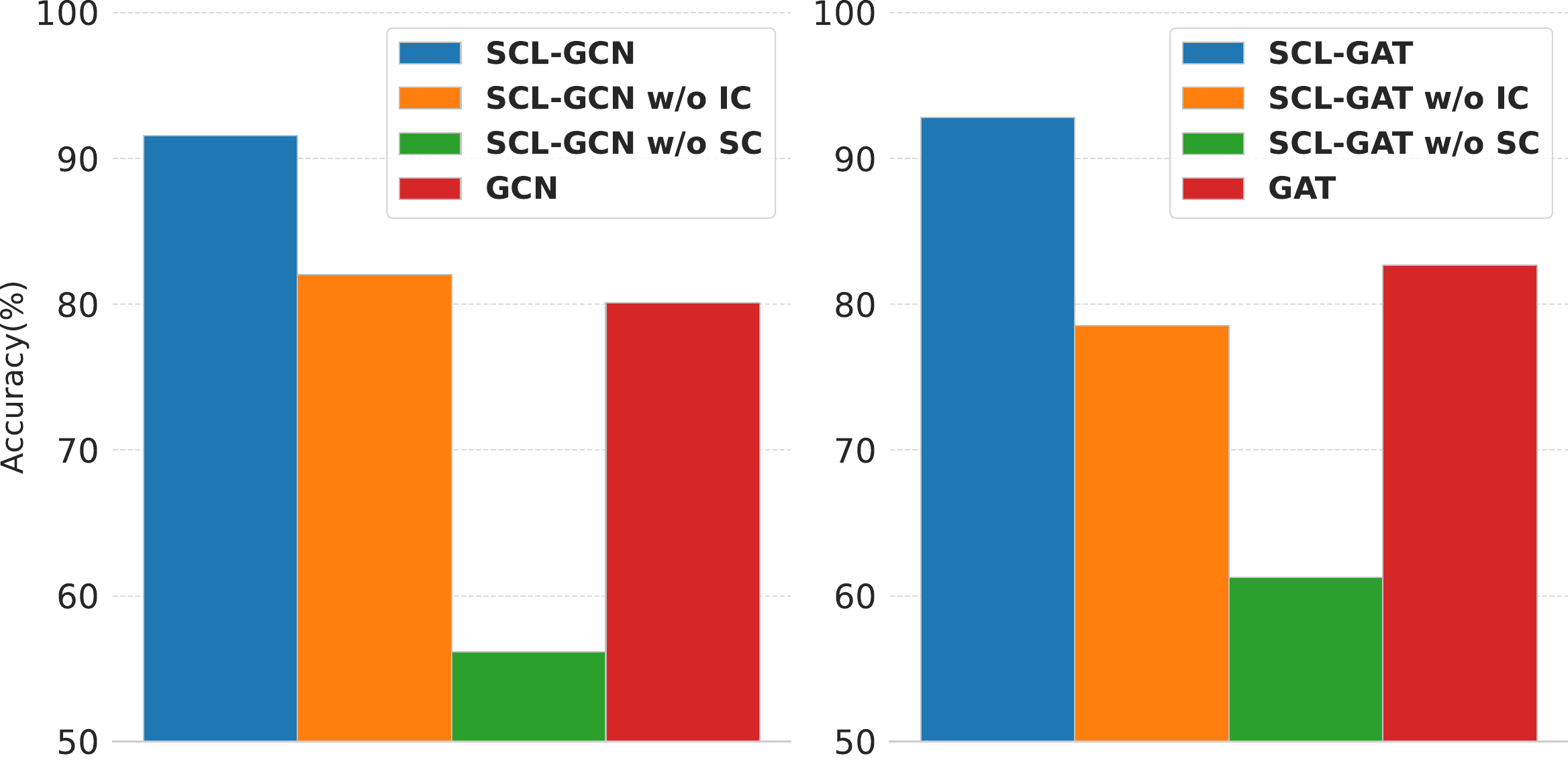}
    \caption{Ablation study of SCL-GNN with GCN and GAT.}
    \label{fig:4}
\end{figure}

\noindent\textbf{Ablation Study on Correlations} 
To further verify the effectiveness of the spurious correlation learner, we conduct another ablation study to understand the correlation effects on performance. We compare SCL-GNN on the same OOD portion of \texttt{Pubmed} with the same training set ratio with: \textbf{SCL-GNNw/oIC:} SCL-GNN without all the irrelevant correlations with normalized values greater than 0.5. \textbf{SCL-GNNw/oSC:} SCL-GNN without all the significant correlations with normalized values smaller than 0.5. Results in Fig.\ref{fig:4} show SCL-GNN achieves the best performance, indicating each learned correlation contributes to effectiveness and robustness and complements each other.

\subsection{Optimization Effect}
\begin{figure}[ht]
\centering
\begin{minipage}[t]{\linewidth}
\subfigure[Bi-level optimization]{
\begin{minipage}[t]{0.5\linewidth}
\centering
\includegraphics[width=\textwidth]{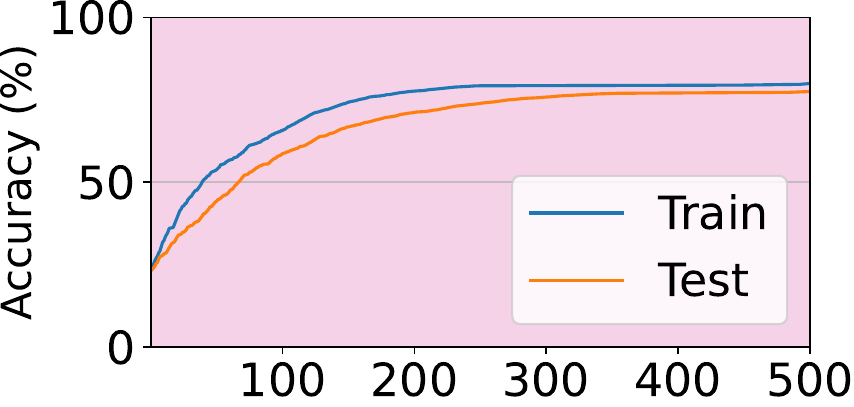}
\end{minipage}%
}%
\subfigure[$\lambda = 1$]{
\begin{minipage}[t]{0.5\linewidth}
\centering
\includegraphics[width=\textwidth]{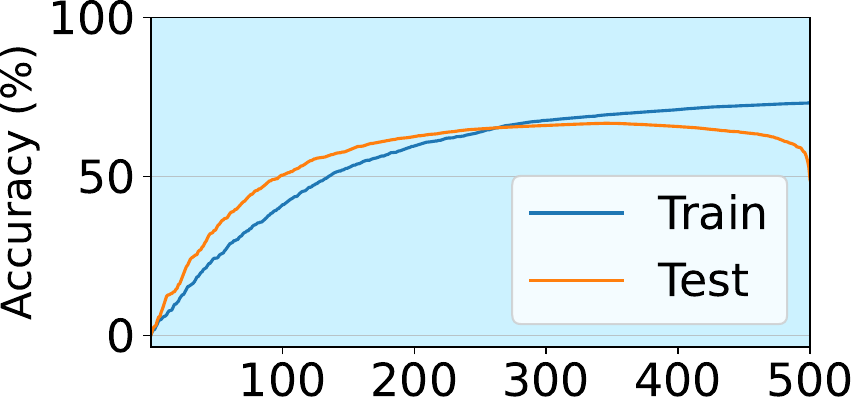}
\end{minipage}%
}%

\subfigure[$\lambda = 2$]{
\begin{minipage}[t]{0.5\linewidth}
\centering
\includegraphics[width=\textwidth]{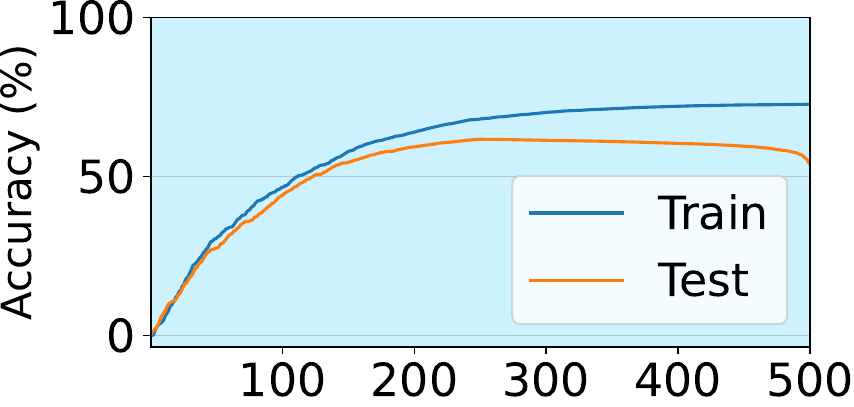}
\end{minipage}%
}%
\subfigure[$\lambda = 3$]{
\begin{minipage}[t]{0.5\linewidth}
\centering
\includegraphics[width=\textwidth]{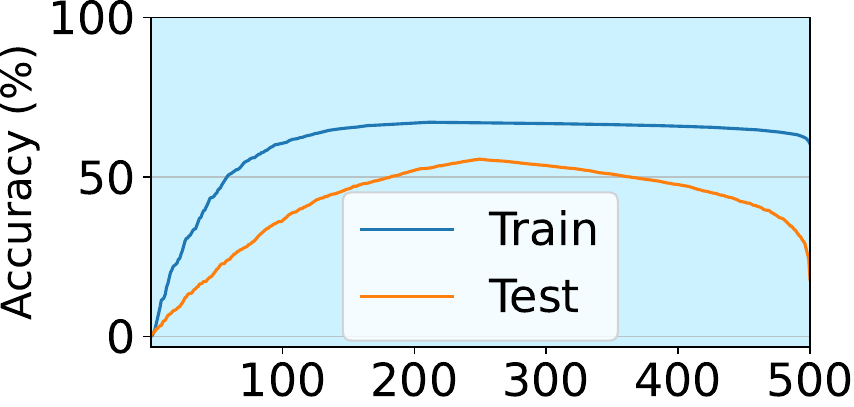}
\end{minipage}%
}%
\end{minipage}
\caption{Comparison of a GCN backbone's learning curves}
\label{fig:5}
\end{figure}
\noindent To answer \textbf{RQ 3}, we reformulate Problem~\ref{problem1} as an optimization problem with a regularization term, where the hyperparameter $\lambda$ controls the weight of the aligner module's learning loss during training. As illustrated in Figure~\ref{fig:5}, the results demonstrate two key findings: (i) Bi-level optimization aligns test accuracy closely with training accuracy, and (ii) non-two-layer optimization accelerates model convergence but leads to significant performance degradation, primarily due to overfitting and reduced generalization. These findings validate the effectiveness of our bi-level optimization formulation.

\subsection{Mechanism Study}
\begin{figure}[ht]
\centering 
\begin{minipage}[t]{\linewidth}
\subfigure[\texttt{Cora}]{
\begin{minipage}[t]{0.5\linewidth}
\centering
\includegraphics[width=\textwidth]{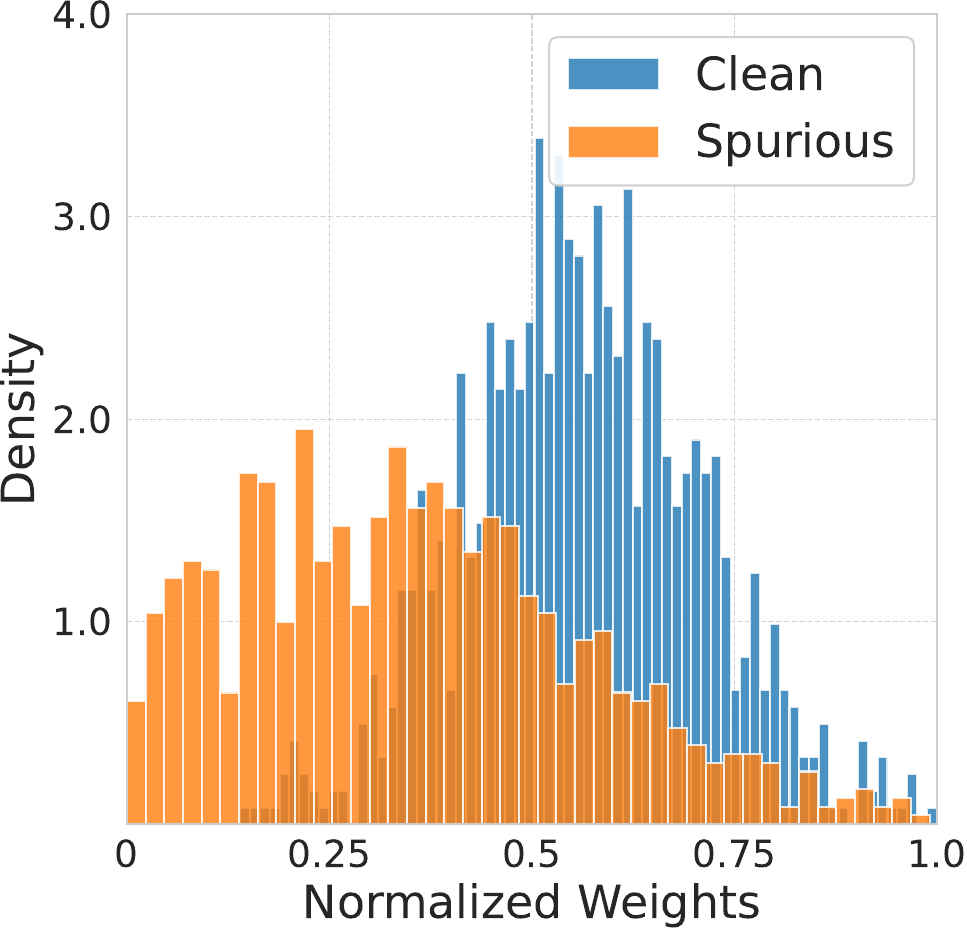}
\end{minipage}%
}%
\subfigure[\texttt{Arxiv}]{
\begin{minipage}[t]{0.5\linewidth}
\centering
\includegraphics[width=\textwidth]{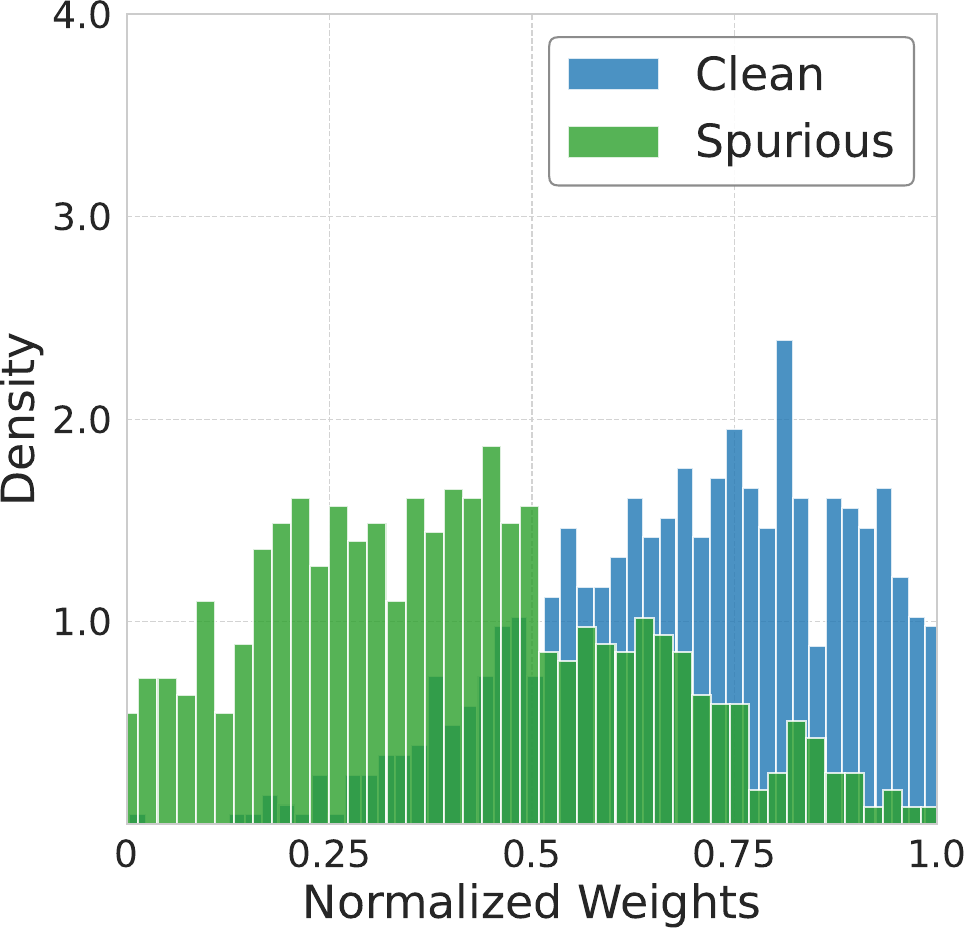}
\end{minipage}%
}%
\end{minipage}
\caption{The distribution of two sets of correlations.}
\label{fig:6}
\end{figure}
To answer \textbf{RQ 4}, we study the spurious learning mechanism's contribution to generalizable GNN under distribution shifts via experiments on \texttt{Cora} and \texttt{Arxiv} datasets. We visualized the fine-tuned weight matrix of the GCN backbone model, as shown in Fig.\ref{fig:6}. The median weight in the spurious node features is lower than that in the clean node features, indicating that SCL-GNN can identify the spurious correlations between labels and features. The weight variance of spurious node features is larger than in clean, showing SCL-GNN's reliable mitigation of spurious correlations.
\section{Conclusion and Discussion}
This work addresses the challenge of enhancing the generalization of GNNs in the presence of spurious correlations, especially under unobserved distribution shifts. By combining theoretical analysis with practical intuition, we use GNN explanation methods to quantify the relevance of node features to prediction labels. Based on these insights, we introduce a learning objective designed to identify and mitigate spurious correlations through the fine-tuning of pre-trained GNN models. Our experiments across four real-world datasets with varying distribution shifts demonstrate the effectiveness of this framework.

While this approach improves the development of generalizable GNNs, several directions remain for future research. The proposed methodology could be applied to tasks such as molecular property prediction~\cite{hao2020asgn}. Furthermore, the principles of spurious correlation learning might be adapted for OOD detection~\cite{li2022ood} to better identify shifts in data distributions.SCL
\bibliographystyle{named}
\bibliography{ijcai25}

\end{document}